\documentclass[pre,aps,twocolumn,floatfix,superscriptaddress]{revtex4-2}
\usepackage{amsmath}
\usepackage{amssymb}
\usepackage{graphicx}
\usepackage{color}
\usepackage{amssymb}
\usepackage{soul}
\usepackage[utf8]{inputenc}
\usepackage[T1]{fontenc}
\usepackage{array}
\usepackage{ulem}

\begin{document}

\title{Reservoir Computing Using Complex Systems}

\author{N. Rasha Shanaz}
\affiliation{Department of Physics, Bharathidasan University, Tiruchirappalli 620024, India}

\author{K. Murali}
\affiliation{Department of Physics, Anna University, Chennai 600025, India}
\author{P. Muruganandam}
\affiliation{Department of Physics, Bharathidasan University, Tiruchirappalli 620024, India}

\begin{abstract}
Reservoir Computing is an emerging machine learning framework which is a versatile option for utilising physical systems for computation. In this paper, we demonstrate how a single node reservoir, made of a simple electronic circuit, can be employed for computation and explore the available options to improve the computational capability of the physical reservoirs. We build a reservoir computing system using a memristive chaotic oscillator as the reservoir. We choose two of the available hyperparameters to find the optimal working regime for the reservoir, resulting in two reservoir versions. We compare the performance of both the reservoirs in a set of three non-temporal tasks: approximating two non-chaotic polynomials and a chaotic trajectory of the Lorenz time series. We also demonstrate how the dynamics of the physical system plays a direct role in the reservoir's hyperparameters and hence in the reservoir's prediction ability. 
\end{abstract}


\flushbottom

\maketitle
\section{Introduction}
	
The emergence of chaos theory showed that the discrepancies in predicting and forecasting highly nonlinear systems, like weather and economy, are due to the chaos inherent in them. For years, this sensitivity to initial conditions hindered long term predictions of complex systems. With the recent developments in machine learning and artificial neural networks, forecasting nonlinear chaotic data is becoming plausible and more accurate. These computing paradigms show remarkable potential in predicting highly nonlinear data, which was otherwise impossible.  
	
One such paradigm, gaining momentum at present, is Reservoir Computing (RC)~\cite{Jaeger2001}. The RC architecture consists of a \textit{reservoir}, a randomly connected artificial recurrent neural network (RNN), which projects the input into a higher dimensional space and a \textit{readout} layer, which analyses patterns in the reservoir states. The reservoir is kept fixed, and only the readout is trained~\cite{Jaeger2004}. This makes training, computationally inexpensive and the reservoir, highly adaptable. The flexible architecture of RC enables a wide range of physical systems to be used in place of RNNs as computing substrates~\cite{Tanaka2019}. Analog electronic circuits \cite{jensen2018}, discrete dynamical systems \cite{Snyder2013}, optical systems~\cite{Vandoorne2008}, and biological cells \cite{Jones2007} are a few of the diverse examples. The RC method has proven competence in forecasting spatiotemporally chaotic dynamics, in the absence of defining models and only with observational data for the system’s evolution~\cite{Pathak2018}. 

The inherent dynamic nature of the RC paradigm renders itself an appropriate candidate for intrinsically dynamic tasks. This paper aims to integrate the model-free prediction ability of reservoir computing with the intrinsic computational potential of complex dynamical systems. We computationally implement a reservoir using a chaotic memristive circuit and study its efficiency in approximating two nonlinear, non-chaotic polynomials and the chaotic Lorenz time series. From the available hyperparameters, we built two reservoirs and analysed how the corresponding dynamics play a role in the efficiency of approximation. We find that the reservoir performs exceptionally well in approximating non-chaotic data but forecasting the chaotic time series requires drastic modification in the reservoir parameters.   
	
\subsection{Reservoir Computing architecture}

\begin{figure}[!ht]
	\centering\includegraphics[width=\columnwidth, clip=true, trim=20mm 35mm 30mm 40mm ]{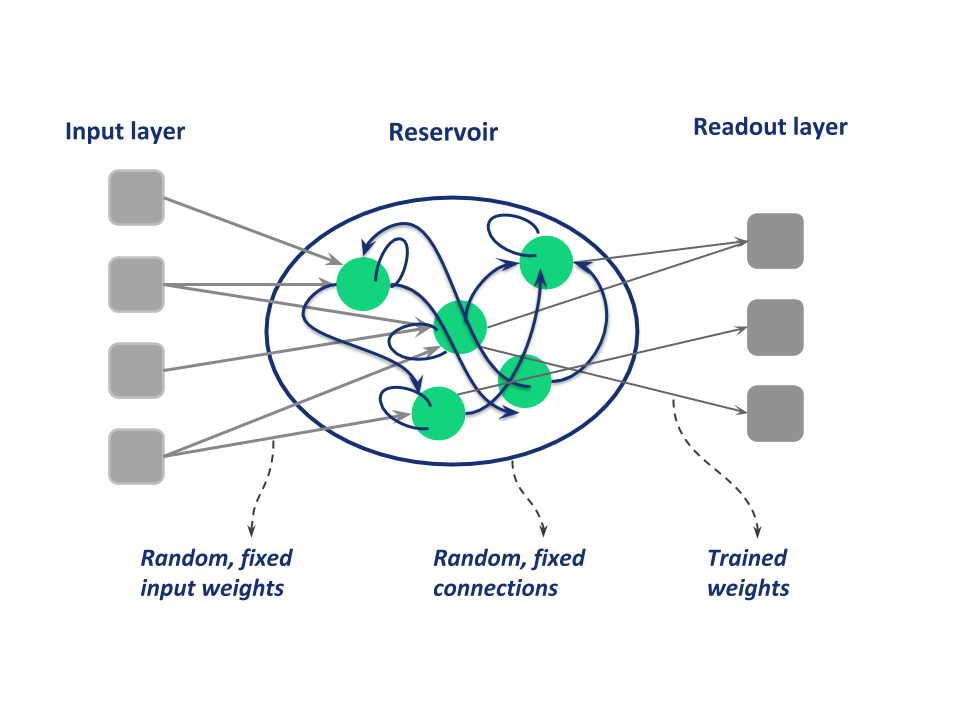}
	\caption{Schematic diagram of a general reservoir computing structure.}
	\label{fig:RC}
\end{figure}

The basic framework of a reservoir computing system consists of an input layer, the reservoir, and the output or readout layer, as shown in Figure~\ref{fig:RC}. %
Conventionally, the reservoir is a randomly connected RNN. But any dynamical system with a potential to dynamically respond to inputs can function as a reservoir~\cite{Tanaka2019}. 
	
The reservoir converts the inputs given at the input layer into a spatiotemporal pattern. The recurrent connections in the reservoir bring  about an intrinsic memory effect and nonlinearly map the input into a higher-dimensional space. This transformation of input into higher-dimensional space resembles what a complex dynamical system could do, hence introducing the possibility of utilising complex systems themselves as reservoirs. The reservoir states are the dynamical response of the reservoir system, which are then mapped to the readout layer. The readout layer is trained using a supervised, non-temporal mapping task like a linear regression~\cite{Lukosevicius2009}. The reservoir offers a sufficiently rich cache of responses for an array of tasks. This rules out the necessity to train the reservoir at every stage and for each different task. 
	
\subsection{Optimal operating regime}

For the reservoir system to efficiently perform computational tasks it should have a high dimensionality for the mapping inputs with reservoir states, and nonlinearity for extracting nonlinear dependencies of inputs. It should have fading memory, also known as the echo state property, which means that the influence of past inputs on the current reservoir states and outputs asymptotically fades out~\cite{Jaeger2001, Maass2002}. The mathematical definition says that a system has echo state property if different initial states converge~\cite{Boedecker2012}. Along with these, it should have separation property, which maps different inputs onto different reservoir states, and approximation property, which maps infinitesimally different reservoir states onto identical targets.

Traditionally, this is realised through a recurrent network of numerous reservoir nodes, but can also be achieved by time-multiplexing the output of a single nonlinear node. This makes the state space infinite dimensional, but the dynamics of the system remains finite dimensional~\cite{Appeltant2011}.

From the complex system studies point of view, the reservoir can be regarded as a complex dynamical system that operates optimally in a certain dynamical regime. When a reservoir system functions in stable dynamical regions, small variations in initial conditions will not reflect current dynamics, reducing the complexity of the model and increasing the bias. This might result in \textit{underfitting} of data. Whereas, when it functions in chaotic regimes, sensitivity increases and amplifies infinitesimal variations, compromising the approximation of closer values and hence increasing variance. This could cause \textit{overfitting} and the reservoir will not be versatile for different data sets. If the reservoir is operated in a dynamical region, where the dynamics are neither too contracting nor too diverging, it will be able to pick up variations in data and also bring about a sweeping approximation of hypotheses. Such an optimal working regime can be obtained at the \textit{edge of chaos}, a region where both order and chaos coexist. ~\cite{Legenstein2007,Inubushi2017}.
	
In the wide variety of available nonlinear dynamical systems, the motive is to choose a simple system that is capable of exhibiting rich nonlinear dynamics and has options for more hyperparameters allowing more precision. A system with chaotic dynamics is an ideal choice for a reservoir because it easily fulfils the above mentioned criteria. 


\section{Methods}
	
\subsection{The reservoir}
\label{sec:2.1}
	
Based on the above mentioned properties of the reservoir, we chose a chaotic oscillator circuit with a memristive element.  %
\begin{figure}[!ht]
\centering\includegraphics[width=\columnwidth]{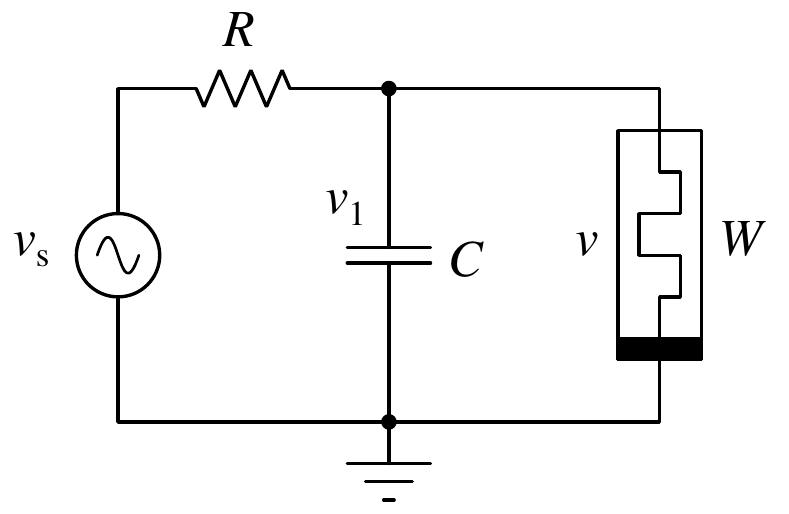}
\caption{The second order memristive oscillator used as the reservoir}
\label{fig:ckt}
\end{figure}%
It is a second order non-autonomous memristive oscillator (Figure \ref{fig:ckt}), built by modifying the memristive Chua's circuit~\cite{Xu2017}. We took into consideration, its simple circuit architecture, experimental reproducibility, rich dynamics,and the inherent fading memory property of the memristor \cite{Ascoli2016a}.%
\begin{figure*}[!ht]
\centering\includegraphics[width=\linewidth]{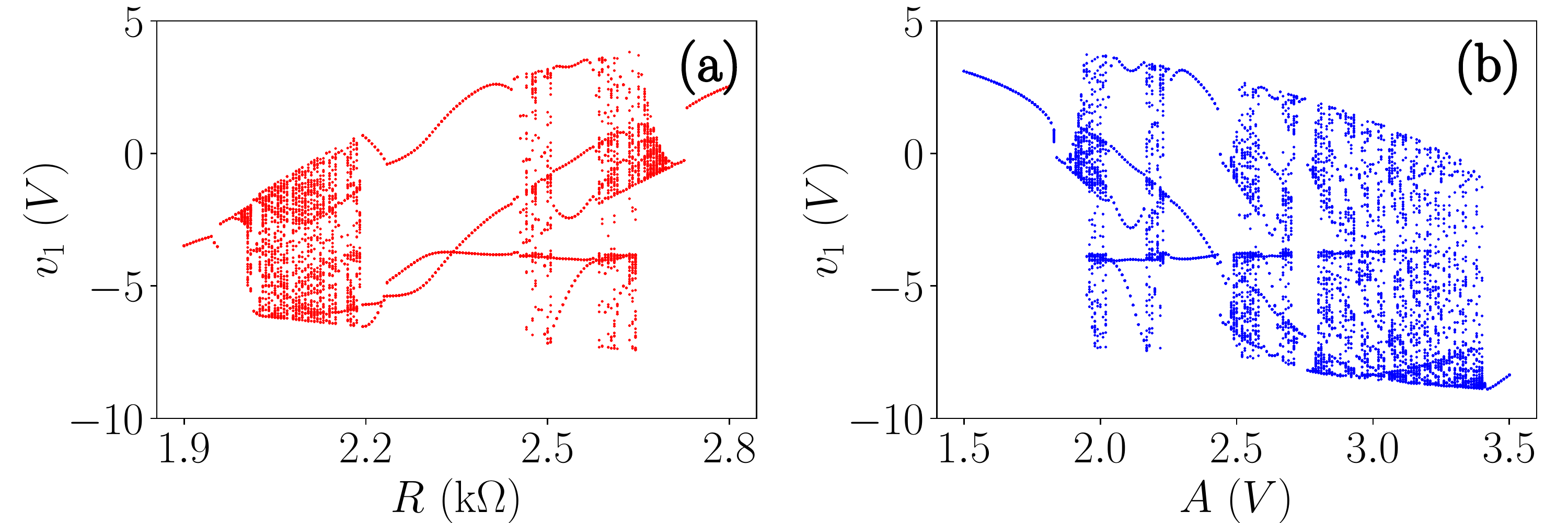}
\caption{Bifurcation diagrams of $ v_1$ with (a) increasing resistance $R$ and (b) increasing forcing amplitude $A$.}

\label{fig:bif}
\end{figure*}%

The governing equations of the circuit are: %
\begin{subequations}%
\label{eq:ckt}
\begin{align}%
\dfrac{dv_{1}}{dt} & = \dfrac{v_{s}-v_{1}}{RC_{1}} +\dfrac{(1-gv_{0}^{2})v_1}{R_{3}C_{1}} , \\
\dfrac{dv_{0}}{dt} & = - \dfrac{v_{1}}{R_{1}C_{0}} -\dfrac{v_{0}}{R_{2}C_{0}} ,
\end{align}%
\end{subequations}%
where $v_{s}= A \sin(2 \pi \nu t)$ is the forcing signal with amplitude $ A $ and angular frequency $ \omega=2 \pi \nu $. The memristor ($W)$ is modeled based on \cite{Xu2016a} as
\begin{align}
i = W(v_0)v = -\dfrac{1}{R_{3}}  \left (1-gv_{0}^{2} \right ) v,
\end{align}
where $ W(v_0) $, $ i $ and $ v $ are the memductance, current and voltage of the memristor, respectively, and $ g $ is the internal gain factor. For computational convenience, we use the following set of normalized version of the governing equations. 
\begin{subequations}
\label{eq:model}
\begin{align}	
\dot{x} & = -\alpha y - x, \\
\dot{y} & = -\beta \left (A' \sin \omega' t  -y \right) + \gamma \left(1-x^{2} \right)y,
\end{align}
\end{subequations}%
where  
$ v_{0} = x / \sqrt{g}$, 
$ v_{1} = y / \sqrt{g}$,
$ A = {A'} / {\sqrt{g}}  $,
$ t = R_{2} C_{0} t' $, 
$ \alpha = {R_{2}}/{R_1} $,
$ \beta = {R_2 C_0}/{RC_1} $,
$ \gamma = {R_2C_0}/{(R_3C_1)} $, and
$ \omega = {\omega'} / {(R_2C_0)}$.
Here, $x$ corresponds to the dimensionless voltage across the capacitor $C$ and $y$ corresponds to the dimensionless  voltage through the memristor $W$. The term $A \sin \omega t$ is the external forcing signal with amplitude $A$ and angular frequency $\omega$. The values of the memristor parameters were set as $R_1 = 8.0~\mbox{k}\Omega$, $R_2 = 4.0~\mbox{k}\Omega$, $R_3 = 1.4~\mbox{k}\Omega$, $C_0 = 4.7~nF$, $C_1 = 6.8~nF$ and $g = 0.1$~\cite{Xu2017}.

The dynamics of the circuit is governed by forcing amplitude ($A$), resistance ($R$), and forcing frequency ($\nu$). The availability of more parameters is advantageous for fine tuning the reservoir computing abilities of the circuit. Keeping in mind the experimental hardships, we chose $ R $ and $ A $ to be the hyperparameters for the RC. Figure~\ref{fig:bif}(a) shows the bifurcation diagrams of $ v_1 $ with $R$ increasing from $1.9~\mbox{k}\Omega $ to $2.8~\mbox{k}\Omega$, with $A$ fixed at $2~V$, and figure~\ref{fig:bif}(b) shows the bifurcation of $v_1$  with $A$ increasing from $ 1.5~V$ to $3.5~V $, keeping $R$ fixed at $2.6~\mbox{k}\Omega$.

\begin{figure}[htp]
\centering\includegraphics[width=\columnwidth]{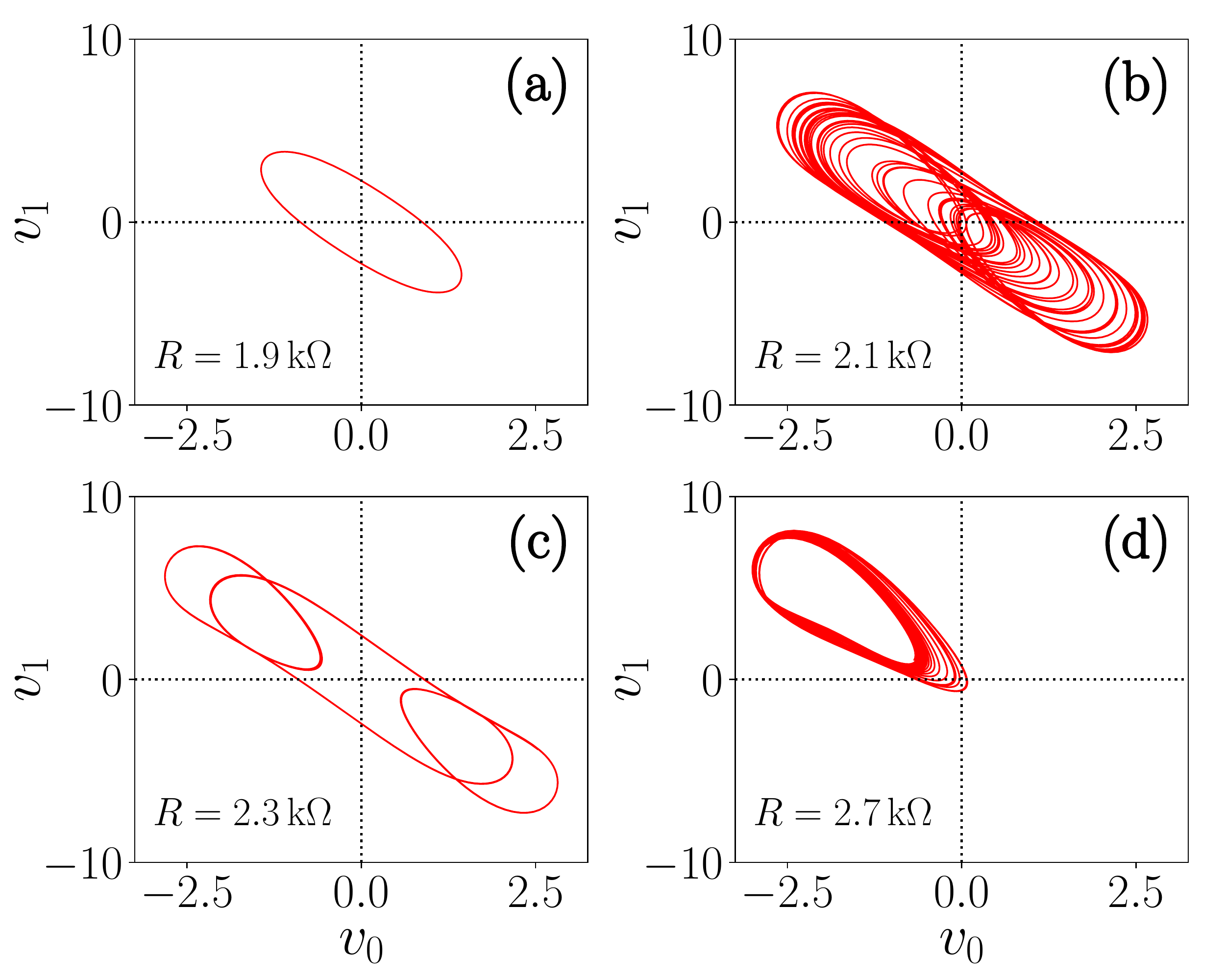}	
\caption{The numerical phase portraits for different values of $R$ on the $ v_0 - v_1$ plane. (a) Period-$T$ limit cycle at $R=1.9\mbox{k}\Omega$. (b) Chaotic attractor at $R=2.1\mbox{k}\Omega$. (c) Period-$3T$ limit cycle at $R=2.3\mbox{k}\Omega$. (d) Chaotic spiral attractor at $R=2.7\mbox{k}\Omega$.}
\label{fig:attrac}
\end{figure}

A wide range of striking dynamical behaviour can be observed, including periodic limit cycles, period doubling bifurcations and chaos. A few selected dynamics are presented by the numerically simulated trajectories of (\ref{eq:ckt}) on the $ v_0-v_1 $ plane, for arbitrary values of $R$, as shown in figure~\ref{fig:attrac}. %
The dynamics corresponding to $A$ have similar features.
Since such environment paves way for the reservoir to be operated in the edge of chaos, the existence of chaotic and periodic windows is essential for this system to be employed as a reservoir.

To analyse the effect of the two hyperparameters on the computational ability of RC, we built two reservoirs: Resistance controlled ($R$-Reservoir) and Amplitude controlled ($A$-Reservoir) versions. For nonchaotic, nonlinear approximating task, we chose the working regime of the $R$-reservoir to be 2.2~$ \mbox{k} \Omega - 2.8~\mbox{k} \Omega $ and that of $A$-reservoir to be $2.1 V - 2.5 V$, which are regions with distinct yet coexisting chaotic and stable dynamics. And for the case of approximating chaotic data, we operated the $R$-reservoir between $2.10~\mbox{k} \Omega - 2.19~\mbox{k} \Omega $, and the $A$-reservoir between $2.75 V - 3.25 V$, which are regions with predominantly chaotic dynamics. The reasoning behind this choice will be explained in section \ref{section3}

\subsection{The task}

We evaluate the performance of the reservoir on a non-temporal task of approximating three different data sets of varying complexity. We estimate the outcome for a new data set based on the training on a known data set. The three data sets used for the task are:
\begin{itemize}
\item $5^{th}$-degree polynomial:
\begin{align}
f(x) = x^5 - 5x^4 + 5x^3 + 5x^2 - 6x - 1
\label{eq:5deg}
\end{align}
with $x$ ranging from $ -1.25 $ to $ 3.25$
		
\item $9^{th}$-degree polynomial:
\begin{align}
f(x) = &\, x^9 + 3x^8 - 4.5x^7 - 21x^6 + x^5  \notag \\ &\, 
+ 44x^4 + 13x^3 - 25x^2 - 11x,
\label{eq:9deg}
\end{align}
with $x$ ranging from $  -1.791 $ to $ 1.834$
		
\item The Lorenz time series~\cite{Lorenz1963}:
\begin{subequations}
\label{eq:Lor}
\begin{align}	
\dot{x} & = \sigma( y - x),\label{eq:Lora} \\
\dot{y} & = \rho x-y-xz, \\
\dot{z} & = xy-\beta z,
\end{align}
\end{subequations}%
where $ \sigma=10$,  $\rho =28$ and $\beta =2.667 $, with the initial condition set at $ x_0=0.5$, $y_0=1$, and $ z_0=2$. 
We approximate 10,000 data points of the trajectory $x(t)$, leaving sufficient points as transients.
	
\end{itemize}

\subsection{Mapping}
	
The prime concern is to define computation in terms of dynamics of the system. We employ a general framework based on the reservoir computing prediction methods proposed by Jaeger and Haas~\cite{Jaeger2004}. For an input vector $u(t)$, the reservoir state vector is modelled as $r(t) = F[ W_{in}~u(t) + M\, r(t-1) ]$, where $t$ denotes discrete time, $W_{in}$ is the input weight matrix, $M$ is the reservoir weight matrix and $F$ is the activation function, which can be any differentiable function and its form may not necessarily be known ~\cite{Lu2018}. Here the discrete-time parameter $t$ does not represent time in a real sense;  instead, it represents the parameter on which the input vector $u(t)$ depends. The above equation represents a non-autonomous dynamical system driven by an external input $u(t)$~\cite{Tanaka2019}. The output vector is given by $y(t) = W_{out} r(t)$, where, $W_{out}$ is the weight matrix of the readout, which is trained using ridge regression. The reservoir is considered as a black box model of the system employed~\cite{Jaeger2004}.

We replace the standard neuron models for $ F $,with the circuit's equations (\ref{eq:model}). We apply the input through the parameter $ \beta $ for the $R$-Reservoir and the external forcing signal $A \sin \omega t$ for the $A$-reservoir. The reservoir input parameter $X$ is normalised and mapped to the input range $x$ of the polynomial. Here $X$ can be $R$ or $A$ in the respective reservoir versions and is taken in a linear range $ X_{min} $ to $ X_{max} $, which are the boundary values of the chosen working regime, as specified in section~\ref{sec:2.1}. For instance, for the $R$-reservoir in the task of approximating the $5^{th}$-degree polynomial (\ref{eq:5deg}), the input is normalised by mapping $[R_{min}, R_{max}]$ to $u(t)$ as $[2.1\,\mbox{k} \Omega, 2.19\,\mbox{k} \Omega] \rightarrow [0,1]$. This normalised input $u(t)$ is then mapped to the independent variable, $x$, of the polynomial as $[0,1] \rightarrow [-1.25, 3.25]$. The interval is divided into $10000$ equally spaced points.

For each input $u$, the reservoir state vector is perturbed for $N=5$ periods and for each cycle of the perturbing signal, $k=10$ points are chosen, to generate $kN$ discrete outputs. This gives $kN=50$ outputs for every parameter value. %
\begin{figure}[htp]
	\centering\includegraphics[width=\columnwidth]{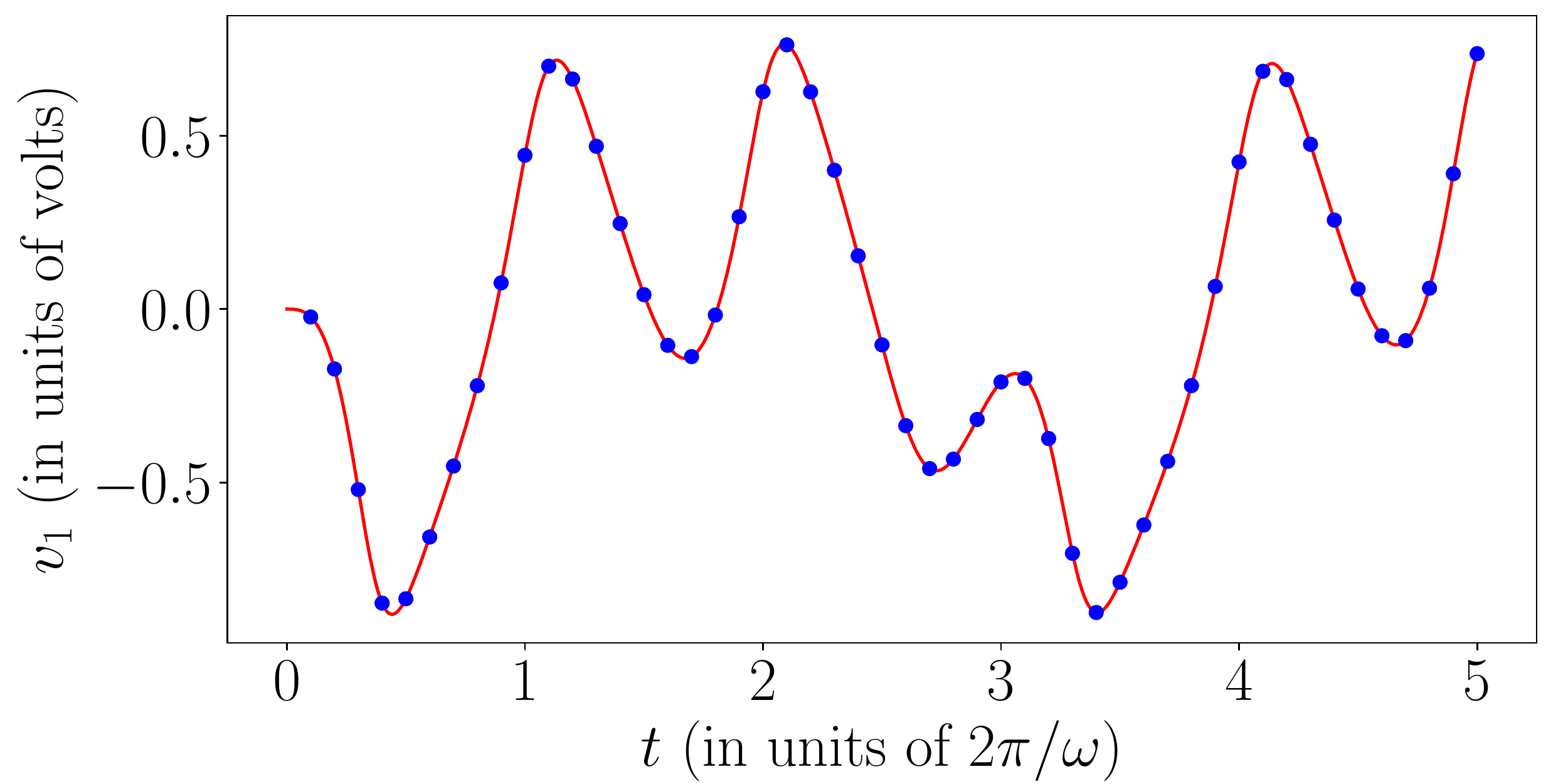}	
	\caption{The $kN=50$ points (blue circles) chosen from the reservoir system output for $N=5$ periods of forcing signal. This output is taken for $R=2.2~\mbox{k}\Omega.$}
	\label{fig:points}
\end{figure}%
\begin{figure*}[!htb]
\centering\includegraphics[width=\linewidth]{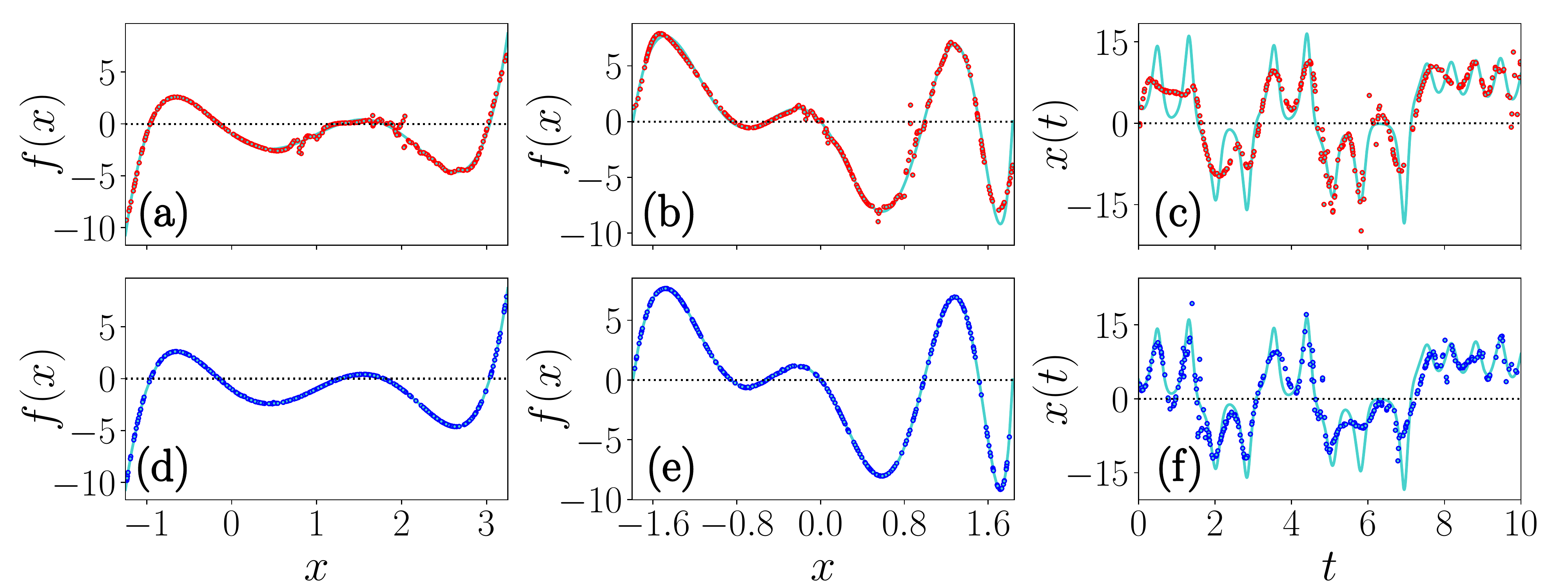}
\caption{The approximated polynomials by the $R$-reservoir (above, in red)  and $A$-reservoir (below, in blue). The target data is shown in cyan. Approximation of (a) the $5^{th}$-degree polynomial (\ref{eq:5deg}) (MSE=$0.0766$), (b) the $9^{th}$-degree polynomial (\ref{eq:9deg}) (MSE=$0.2499$) and  (c) the Lorenz time series (\ref{eq:Lor} (MSE=$11.1691$), by  the $R$-reservoir. Approximation of (d) the $5^{th}$-degree polynomial (MSE=$0.000102$), (e)  the $9^{th}$-degree polynomial (MSE=$0.001170$) and  d(f) the Lorenz time series (MSE=$9.0455$) by  the $A$-reservoir.}
\label{fig:result}
\end{figure*}%
The system is reset to start in the same initial condition between each input. A visual representation of how the $kN$ points are chosen is shown in Figure \ref{fig:points}. Since these points are outputs of a chaotic system, there will be an inherent complexity in them. This brings about the mapping of input to a higher dimensional space, by generating numerous nonlinear transformations as a function of the input parameter $X$, without the requirement for a system dynamics with higher dimensionality. %

We use the Python based Scikit-learn package for ridge regression to train the reservoir for these tasks~\cite{Scikit-learn2011}. We assess the performance of the reservoir using Mean Squared Error (MSE) from cross-validation. The details are outlined in \ref{sec:app:a} 
We optimise numerically using the train and test scheme to test the goodness of the reservoir. We split the data set in a training:testing ratio of $67:33$. In order for the algorithm to learn the overall pattern of the data, the training points are chosen randomly from the entire data range. And the remaining data points are used for the approximation task. We also study the effect of the working parameter range of the reservoir on its bias and variance and thereby analysing the best working regime for approximation tasks. 

\section{Results and discussion}
\label{section3}

The three data sets were approximated by the reservoir and the approximated curves of the $5^{th}$- and $9^{th}$-degree polynomials, and the Lorenz time series are shown in Figure \ref{fig:result}. All points shown on the plot are the approximated values and the training data are not shown here. For comparison, the target data is presented. For all three cases, the reservoir system parameters were kept fixed. The goodness of fit is assessed through the loss function, Mean Squared Error (MSE).%
\begin{figure*}[!htb]
\centering\includegraphics[width=\linewidth]{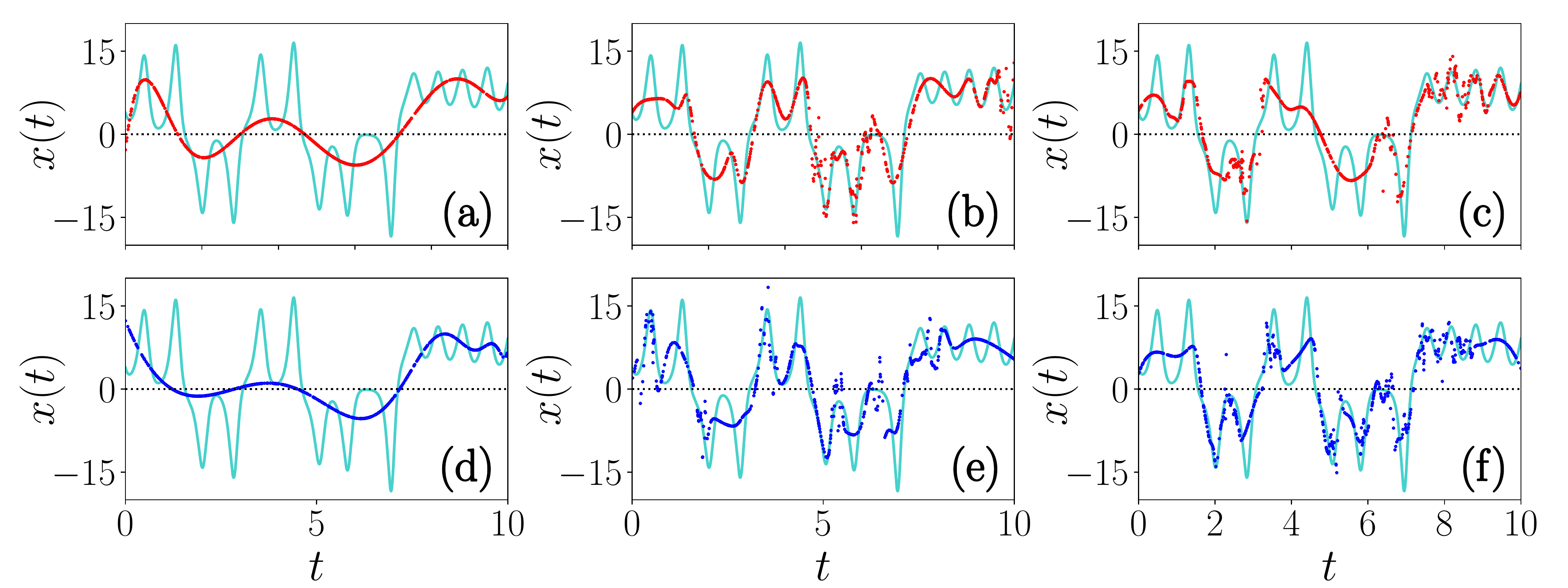}
\caption{The various approximations of the Lorenz time series by the reservoirs in different dynamical regions, demonstrating the effect of the system dynamics on computational capability of the reservoir.  The plots show approximation by the $R$--reservoir (above, in red) operated in (a) periodic regime ($ R= 2.3\mbox{k}\Omega - 2.4\mbox{k}\Omega $ ) (b), chaotic regime $ R=2.10\mbox{k}\Omega - 2.19\mbox{k}\Omega$ and (c) the full bifurcation parameter range ($ R = 1.9\mbox{k}\Omega - 2.8\mbox{k}\Omega$), and the approximation by the $A$--reservoir (below, in blue) in (d) periodic regime ($A= 1.5V - 1.8V $), (e) chaotic regime ($ A= 2.5V-3.4V $) and (f) the full bifurcation parameter range ($ A= 1.5V-3.5V $).}
\label{fig:result2}
\end{figure*}
The variances of the reservoirs were regulated to minimise the loss function and to accommodate the increasing complexity in the three data sets and thereby, minimising error. 

The performance of the reservoir, in terms of approximation, is good in the case of the $5^{th}$- and $9^{th}$-degree polynomials. Comparing the mean square errors for the two reservoirs, the $A$-reservoir has a very low value of the order $10^{-4}$ for $5^{th}$-degree and $ 10^{-3}$ for the $9^{th}$-degree polynomial. This is two orders of magnitude less than the MSEs of the $R$-reservoir, which are $ 10^{-2}$ and $ 10^{-1}$ for the $5^{th}$- and $9^{th}$-degree polynomials. The absence of noise in the approximation and the low values of MSE show that the $A$-reservoir is better performing. This can be attributed to the fact that the dynamical regime, in which the A-reservoir was operated in ($ 2.0 V - 2.5 V$) has a good balance of order and chaos, and thus bringing the optimal balance between bias and variance.
	
Now, in the case of Lorenz time series, we plot 10,000 data points of the time series for $10$ units of time. In the approximation tasks, the MSEs are too high but the overall shape of the approximated polynomial resembles the time series, for both the reservoirs. Improving the quality of approximation, in this case, will require increasing the variance to very high values in order to attune to the chaos. Such an increment will result in the reservoir capturing the minute variations along with the underlying pattern. This will lead to overfitting of other data sets, hence compromising the long term prediction ability of the reservoir. This led us to study the effect of different dynamical regions on the computational ability and thereby eliciting the optimal working regime for predicting chaotic data.

Taking the same time series, we trained both the A-controlled and $R$-controlled reservoirs in three different dynamical regimes of the memristive circuit. We chose the regions with only periodic dynamics, only chaotic dynamics, and the entire bifurcation range. Figure \ref{fig:result2} shows, the various approximations by the $R$-reservoir and the $A$-reservoir and the corresponding parameter ranges in which they were operated.
	
The reservoir system dynamics for a particular range of hyperparameters ($R$ and $A$), as seen in the bifurcation diagram, is reflected in the corresponding approximation of the Lorenz equation. For instance, for the  Figure \ref{fig:result2}(c) the reservoir parameter $R$ is set in the range $1.9 \mbox{k}\Omega - 2.8 \mbox{k}\Omega$. The bifurcation diagram of $R$ given in Figure \ref{fig:bif}(a) shows alternating periodic and chaotic regions. In the figure, we find that the approximated curve is flatly approximated in ranges corresponding to periodic dynamics and is highly sensitive in ranges corresponding to chaotic dynamics, mirroring the system dynamics. Similar interpretations can be drawn by comparing the bifurcation plot region for a particular parameter range with the corresponding plot of approximation done by the reservoir operating in that range.
	
This demonstrates how system dynamics will alter bias and variance of a reservoir built of a complex system. In the stable regime, a system will behave predictably and has a good fading memory. Small variations in initial conditions or noise will not be reflected in the current dynamics. Hence along with rejecting noise, a reservoir operating in this region will fail to capture the underlying pattern in the data. On contrary, a system in the chaotic regime will have high sensitivity to infinitesimal changes and a diverging behaviour. This will compromise the reservoir's ability to approximate close-lying values and hence rendering  it incapable of predicting other data sets. And hence a system with highly chaotic dynamics will be better able to predict a chaotic data set.
	
\section{Conclusion}

We demonstrated how a single node reservoir, made of a simple memristive chaotic oscillator, can be employed to approximate higher degree polynomials and how its dynamics play a role in the bias and variance of the reservoir's computational abilities. Of the two reservoirs built, the amplitude controlled version performs better due to the optimal working region offered by the parameter. The approximated data matches the target data set with a very small mean squared error. By operating the reservoir in different dynamical regimes, we re-established how chaos and order affect the variance and bias of the reservoir.

This work is limited to non-temporal tasks but holds open the future possibility for temporal time series forecasting. The presence of memristor in the circuit makes it a promising candidate for human neural analysis and studying complex adaptive systems.It is worthy to note that reservoirs made of memristor networks satisfy fading memory and separation properties \cite{Sheldon2020}.

There are many real-world scenarios where order and chaos play an important role, ranging from epileptic seizures to binary star systems. The inherent nonlinearity in these systems makes long term prediction difficult. The current scenario in machine learning and artificial neural networks poses a promising future. Incorporating the dynamics of nonlinear complex systems will take forward the current developments by many steps. 
\appendix

\section{Regression and cross validation}
\label{sec:app:a}
\subsection{Ridge regression}
In machine learning, regression analysis is performed to estimate the relationship between the independent variable and dependant outcome variable(s). It establishes a relationship between the input and output variables of the reservoir and hence helps in prediction or forecasting tasks. The most common method is linear regression where we find a line that most closely fits the data. For a linear model of $i$ number of features $X_i$ given as $Y_i = \underset{i}{\sum} w_i X_i + b$, the regression aims to optimise the slope $w_i$ and the intercept $b$ by minimising the cost function given as 
\begin{align}
\underset{w}{\text{min\ }}\left( \underset{i}{\sum} (w_i X_i - Y_i)^2 \right).
\label{eq:ridge2}
\end{align}
Ridge regression is a type of linear regression which uses regularisation and seeks to minimise the sum of squared error between the model and the training data.  
The ridge coefficients minimise the cost function by imposing a penalty on the coefficient sizes equal to the square of their magnitudes as
\begin{align}
\underset{w}{\text{min\ }} \left( \underset{i}{\sum} \left(w_i X_i - Y_i\right)^2 + \alpha ~\underset{i}{\sum} w_i^2 \right),
\label{eq:ridge4}
\end{align}
where $\alpha \geq 0$ is the complexity parameter. 

\subsection{Cross-validation}
It is a technique for assessing the validity of the model learned by the machine learning algorithm. It is done by partitioning the data into complementary subsets, analysing one \emph{training} subset and validating the analysis on the other \emph{validation} subset. Performing this multiple times will help reduce the variance. The cross-validation estimator will select the best hyperparameter value. 

\section*{Acknowledgement}
	
The work of PM forms parts of research projects by Council of Scientific and Industrial Research (CSIR), India under Grant No. 03(1422)/18/EMR-II, and Science and Engineering Research Board (SERB), India under Grant No. CRG/2019/004059.
	


\end{document}